\documentclass{article}

\usepackage{microtype}
\usepackage{graphicx}
\usepackage{subfigure}
\usepackage{booktabs} 

\usepackage[accepted]{icml2024}
\definecolor{CColor}{rgb}{0.01,0.31,0.59}
\definecolor{GGray}{rgb}{0.80,0.90,1}
\definecolor{Shady}{rgb}{0.9,0.9,0.9}
\definecolor{kaistblue}{RGB}{20,135,200}
\definecolor{kaistdarkblue}{RGB}{0,65,145}
\definecolor{urbanablue}{RGB}{19,41,75}
\definecolor{urbanaorange}{RGB}{232,74,39}
\definecolor{drp}{rgb}{0.53,0.15,0.34}
\usepackage[plainpages=false,pdfpagelabels,colorlinks=true,linkcolor=CColor,citecolor=CColor,urlcolor=CColor]{hyperref}
\usepackage{amsmath}
\usepackage{amssymb}
\usepackage{mathtools}
\usepackage{amsthm,multirow,tcolorbox}

\usepackage[capitalize,noabbrev]{cleveref}

\theoremstyle{plain}

\theoremstyle{definition}

\theoremstyle{remark}

\usepackage{url}
\usepackage{booktabs}
\usepackage{graphicx} 
\usepackage{diagbox}
\usepackage{caption}
\usepackage{colortbl}
\usepackage{subcaption}
\usepackage{resizegather}
\usepackage{pifont}
\newcommand{\cmark}{\ding{51}}%
\newcommand{\xmark}{\ding{55}}%
\newcommand{\mbf}{\mathbf}%
\newcommand{\gr}{\rowcolor[gray]{.95}}
\newcommand*\web{\vcenter{\hbox{\includegraphics[width=1.3em]{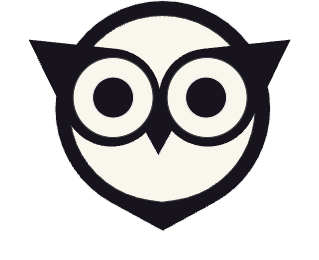}}}}

\newcommand*\circled[1]{\tikz[baseline=(char.base)]{
            \node[shape=circle,draw,inner sep=0.4pt] (char) {#1};}}

\usepackage[textsize=tiny]{todonotes}

\icmltitlerunning{Outlier Weighed Layerwise Sparsity: A Missing Secret Sauce for Pruning LLMs to High Sparsity}

\begin{document}

\twocolumn[
\icmltitle{Outlier Weighed Layerwise Sparsity (OWL$\web$): \\ A Missing Secret Sauce for Pruning LLMs to High Sparsity}




\begin{icmlauthorlist}
\icmlauthor{Lu Yin}{Surrey,TUe,Google}
\icmlauthor{You Wu}{Google}
\icmlauthor{Zhenyu Zhang}{UT}
\icmlauthor{Cheng-Yu Hsieh}{UW}
\icmlauthor{Yaqing Wang}{Google}
\icmlauthor{Yiling Jia}{Google}
\icmlauthor{Gen Li}{CU}
\icmlauthor{Ajay Jaiswal}{UT}
\icmlauthor{Mykola Pechenizkiy}{TUe}
\icmlauthor{Yi Liang}{Google}
\icmlauthor{Michael Bendersky}{Google}
\icmlauthor{Zhangyang Wang}{UT}
\icmlauthor{Shiwei Liu}{ox,TUe}
\end{icmlauthorlist}
\icmlaffiliation{TUe}{Eindhoven University of Technology, the Netherlands}
\icmlaffiliation{Surrey}{University of Surrey}
\icmlaffiliation{Google}{Google Research, USA}
\icmlaffiliation{UT}{University of Texas at Austin, USA}
\icmlaffiliation{CU}{Clemson University, USA}
\icmlaffiliation{UW}{University of Washington, USA}
\icmlaffiliation{ox}{University of Oxford, UK}

\icmlcorrespondingauthor{Shiwei Liu}{shiwei.liu@maths.ox.ac.uk}

\icmlkeywords{Machine Learning, ICML}

\vskip 0.3in
]



\printAffiliationsAndNotice{\icmlEqualContribution} 

\begin{abstract}
\looseness=-1 Large Language Models (LLMs), renowned for their remarkable performance across diverse domains, present a challenge when it comes to practical deployment due to their colossal model size. In response to this challenge, efforts have been directed toward the application of traditional network pruning techniques to LLMs, uncovering a massive number of parameters that can be pruned in one-shot without hurting performance. Building upon insights gained from previous work, prevailing LLM pruning strategies have consistently adhered to the practice of uniformly pruning all layers at equivalent sparsity, resulting in robust performance. However, this observation stands in contrast to the prevailing trends observed in the field of vision models, where non-uniform layerwise sparsity typically yields stronger results. To understand the underlying reasons for this disparity, we conduct a comprehensive study and discover a strong correlation with the emergence of activation outliers in LLMs, which are output features exhibiting significantly greater magnitudes compared to their counterparts. Inspired by this finding, we introduce a novel LLM pruning methodology that incorporates a tailored set of \textbf{non-uniform layerwise sparsity ratios}, termed as \textbf{O}utlier \textbf{W}eighed \textbf{L}ayerwise sparsity (\textbf{OWL}). The sparsity ratio of OWL is proportional to the outlier ratio observed within each layer, facilitating a more effective alignment between layerwise weight sparsity and outlier ratios. Our empirical evaluation, conducted across the LLaMA-V1 family and OPT, spanning various benchmarks, demonstrates the distinct advantages offered by OWL over previous methods. For instance, OWL exhibits a remarkable performance gain, surpassing the state-of-the-art Wanda and SparseGPT by \textbf{61.22} and \textbf{6.80} perplexity at a high sparsity level of 70\%, respectively, while delivering \textbf{2.6$\times$} end-to-end inference speed-up in the DeepSparse inference engine. Code is available at \url{https://github.com/luuyin/OWL.git}.

\end{abstract}

\section{Introduction}

The remarkable performance exhibited by Large Language Models (LLMs) across a diverse spectrum of applications has ignited an unparalleled race among tech giants and academic institutions to build LLMs at the billion-parameter scale~\citep{brown2020language,touvron2023llama,touvron2023llama2,brown2020language}. The compelling performance of Large Language Models (LLMs) demonstrated in various applications triggers an unprecedented competition of building billion-level LLMs among tech giants and academic institutions~\citep{brown2020language,touvron2023llama,touvron2023llama2,brown2020language}. While their exceptional capabilities are undeniable, the colossal size and computational demands of these models have also raised substantial concerns, particularly in terms of financial expenditure and environment~\citep{luccioni2022estimating,patterson2021carbon}.

\begin{figure*}[ht]
\centering
\centering
\includegraphics[width=0.9\textwidth]{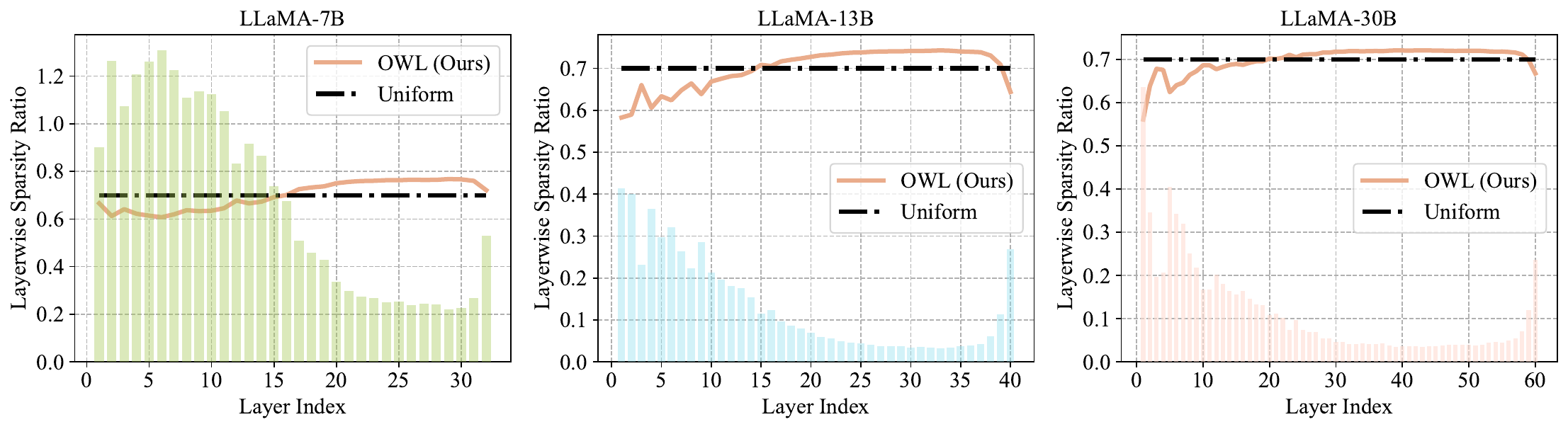} 
\vspace{-0.5em}
\caption{The demonstration of the OWL layerwise sparsity and Uniform layerwise sparsity at 70\% sparsity. The bar chart in the background corresponds to the Layerwise Outlier Distribution (\texttt{LOD}), as elaborated in Section \ref{sec:empirical_study}  } 
\vspace{-0.5em}
\label{fig:layerwise}
\end{figure*}

\looseness=-1 Network pruning~\citep{mozer1989skeletonization,janowsky1989pruning,lecun1989optimal,han2015learning}, as a long-established model compression method, is expected to serve as an effective solution for reducing the size of LLMs. However, network pruning usually favors a certain time of fine-tuning or re-training to reacquire the original optimal performance. Given the extensive text corpus and model size associated with LLMs, conventional fine-tuning becomes exceedingly challenging and less desirable. Fortunately, recent endeavors have explored the possibility of LLM pruning without the need for fine-tuning, showcasing that LLMs contain a substantial number of parameters that can be removed in a single step with minimal performance degradation~\citep{frantar2023massive,sun2023simple,jaiswal2023emergence,ma2023llm}. 
SparseGPT~\citep{frantar2023massive} addresses the challenge of LLM pruning from the perspective of layerwise reconstruction problem.  In this context, the primary goal is to minimize the output discrepancy in terms of the reconstruction error between dense and sparse LLMs. It adopts an iterative strategy to handle the computational hurdle posed by the row-Hessian problem. Specifically, it employs the Optimal Brain Surgeon (OBS) algorithm~\citep{hassibi1993optimal} to selectively prune and update weights in a column-wise manner. 
Wanda~\citep{sun2023simple}, on the other hand, introduces a novel pruning metric that takes into account both the weight magnitudes and their corresponding input activations. Remarkably, it achieves performance on par with SparseGPT without relying on computationally expensive second-order information. The effectiveness of Wanda stems from the emergence of the 
outlier features residing within LLMs. These outliers, which tend to be significantly larger than typical features, are nonetheless crucial for optimizing LLM performance~\citep{kovaleva2021bert,puccetti2022outliers,timkey2021all,dettmers2022llm}. Both SparseGPT and Wanda exhibit appealing performance, showcasing their ability to reduce model parameters by up to 50\% while incurring only a modest increase of  perplexity~\citep{sun2023simple}.

It is worth noting that SparseGPT and Wanda unanimously follow previous work on BERT pruning~\citep{sanh2020movement,kurtic2022optimal} and choose to prune LLMs with a uniform sparsity ratio per layer, \emph{i.e.,} each layer will be pruned at the same sparsity. Such choice is reasonable for LLMs, as the pruning process typically involves sorting the importance scores of weights. Conducting such sorting globally across layers could become a computational bottleneck, especially for models at the billion-parameter scale. Nevertheless, before it has been taken root that uniform layerwise sparsity is the default choice for LLMs, we raise a timely inquiry: \textit{are there any pivotal aspects that have been inadvertently omitted in the context of favorable layerwise sparsity ratios for LLM pruning?} 

Three reasons behoove us to pose the above research question: \underline{\textit{First}}, it is widely acknowledged that within Transformer architectures, certain components hold greater significance than others, and thus, they merit distinct treatment during the pruning process~\citep{wang2020rethinking,bhojanapalli2021leveraging}; \underline{\textit{Second}}, a consensus view has been reached in computer vision that non-uniform layerwise sparsity typically achieves stronger results than uniform sparsity~\citep{liu2022unreasonable,lee2020layer};  \underline{\textit{More importantly}},
LLMs demonstrate astonishingly emergent behaviors~\citep{wei2022emergent,schaeffer2023emergent,dettmers2022llm} as model size continuously scales up, a phenomenon distinct from smaller-scale language models.
These emergent behaviors offer fresh insights into the domain of LLM pruning. For instance, the existence of outlier features within LLMs, with magnitudes up to 20 times larger than others, exerts a profound influence across all Transformer layers~\citep{kovaleva2021bert,dettmers2022llm}.

\textbf{Contributions.} Given the pivotal role that outliers play in LLMs, coupled with the demonstrated effectiveness of Wanda~\citep{sun2023simple}, our initial investigation centers on a systematic examination of the impact of existing LLM pruning methodologies on outliers. We uncover a strong correlation between pruning efficacy and the retention ratio of outliers: contemporary state-of-the-art LLM pruning approaches, such as SparseGPT and Wanda, exhibit remarkable preservation of outliers, even though the former was not originally designed with this intent. Moreover, we conduct an in-depth analysis of the distribution of outliers across different layers and observe  \textbf{a notably non-uniform pattern}. This non-uniform distribution emerges as a valuable indicator for the formulation of layerwise sparsity strategies tailored specifically for LLMs. Building upon this newfound insight, we introduce an LLM pruning paradigm characterized by a novel layerwise sparsity ratio, denoted as \textbf{O}utlier \textbf{W}eighed \textbf{L}ayerwise sparsity (\textbf{OWL}). OWL inherently assigns greater emphasis to layers housing a higher prevalence of outliers, thereby facilitating more nuanced coordination between sparsity in weight matrices and the presence of outliers within the layer.

We conduct extensive experiments to evaluate the performance OWL across a spectrum of LLMs, including LLaMA-V1~\citep{touvron2023llama}, and OPT~\citep{zhang2022opt}, from 7B to 65B. Our empirical results show that OWL consistently outperforms existing top-performing  LLM  pruning methods, particularly at high sparsity levels. For instance,  we observe significant improvements achieved by OWL over Wanda with LLaMa-7B on WikiText~\citep{merity2016pointer}, with perplexity reductions of more than 60 and 3300 perplexity at sparsity levels of 70\% and 80\%, respectively. When evaluated in the DeepSparse \citep{NM} inference engine, OWL delivers a 2.6$\times$ - 3.9$\times$ end-to-end speedup on CPUs with 70\% - 90\% sparsity.

Overall, our research provides a compelling counter-argument to previous studies by highlighting the previously overlooked yet crucial role of layerwise sparsity ratios in LLM pruning. This change in perspective has enabled us to push the boundaries of achievable one-shot LLM pruning ratios to 70\%. \textit{Note that while non-uniform layerwise sparsity has been extensively explored in network pruning, our paper represents the first effort to make it applicable in LLM pruning, challenging the conventional belief that uniform layerwise sparsity is the default and optimal choice for LLM pruning.}

\vspace{-0.5em}
\section{Related Work}

\textbf{Pruning and LLM Pruning.} 
\looseness=-1 Since the 1980s, network pruning has been a well-established technique for simplifying neural networks in various applications while maintaining accuracy~\citep{mozer1989skeletonization,han2015learning,mocanu2018scalable,wen2017learning,lin2019towards}. However, when it comes to pruning Large Language Models (LLMs), progress has been limited. Traditional pruning typically requires a round of re-training to restore performance, which can be challenging for LLMs. To address this challenge, researchers have developed pruning algorithms specifically tailored for LLM compression. For example,~\citet{ma2023llm} explored structured sparse LLMs using Taylor pruning to remove entire weight rows, followed by LoRA fine-tuning~\citep{hu2021lora}. Recent research has shifted toward unstructured pruning without the need for fine-tuning, showing substantial advancements. SparseGPT~\citep{frantar2023massive} utilizes the Hessian inverse for pruning and with subsequent weight updates to reduce reconstruction error of dense and sparse weights, while Wanda~\citep{sun2023simple} produces a criterion incorporating weight magnitude with their input activations, aiming to preserve outlier features. \citet{zhang2023dynamic} extended dynamic sparsity \citep{mocanu2018scalable,evci2020rigging,liu2021we} to efficiently fine-tune sparse LLM without weight updating. Our work for the first time probes the crucial role of non-uniform layerwise sparsity for LLM pruning, making good progress in this field.

\textbf{Layerwise Sparsity for Pruning.} 
While it is common to use uniform layerwise sparsity~\citep{zhu2017prune,gale2019state} to prune language models~\citep{sanh2020movement,kurtic2022optimal}, there is a well-established line of work that explore non-uniform layerwise sparsity in terms of pruning vision models.~\citet{Mocanu2016xbm} propose a non-uniform and scale-free topology inspired from graph theory, showing better performance than the dense counterpart when applied to restricted Boltzmann machines. Follow-up works significantly improve its scalability based on \textit{Erd{\H{o}}s-R{\'e}nyi} graph~\citep{24gotErdos1959}, extending to fully-connected layers~\citep{mocanu2018scalable} and convolutional layers~\citep{evci2020rigging,liu2022unreasonable} as data-free and feedforward-free layerwise sparsity. Another group of work produces non-uniform sparsity by applying a global threshold on every layer~\citep{frankle2018lottery,lee2018snip,wang2020picking,lee2020layer,liu2021sparse}. However, global pruning becomes extremely expensive and inefficacious in the context of LLM pruning as shown in Table~\ref{tab:Lod_sparse_gra}. We also provide a comparison among the most common layerwise sparsity for LLMs in Section~\ref{sec:com_lw_sparsity}, and all of them fail to perform on LLMs.

\textbf{Outliers in LLMs.}
Recent studies have revealed certain emergent characteristics unique to language models at scale. Specifically, one intriguing trait of LLMs is the exhibition of \textit{outlier features}, which are the features with significantly larger magnitudes than others~\citep{kovaleva2021bert,puccetti2022outliers,timkey2021all,dettmers2022llm}.~\citet{kovaleva2021bert} first discovered that a very small number of outlier dimensions appears in the pre-trained encoder layers of GPT-2.~\citet{puccetti2022outliers} further demonstrated that outliers are causally related to high-frequency tokens in pre-training data. While constituting only a very small portion of the entire feature dimensions, these outliers play an imperative role in models' predictive performance. Given the importance of outliers, several recent works have developed techniques to effectively quantize LLMs with minimal performance drop~\citep{dettmers2022llm,xiao2023smoothquant,lin2023awq}. However, in the context of LLM pruning, this unique characteristic has scarcely been explored except for~\citet{sun2023simple}. Our work draws inspiration from the emergent outliers in LLMs, and provides a new technique that leverages the distribution of outliers to guide layerwise LLM pruning.

\vspace{-0.5em}
\section{Outlier Weighed Layerwise Sparsity}
\vspace{-0.5em}

In this section, we will introduce \textbf{O}utlier-\textbf{W}eighed \textbf{L}ayer-wise sparsity (\textbf{OWL}) step by step, from rationale to empirical studies, and eventually to our algorithm.

\subsection{Rationale}

The success of pruning in pre-LLM model compression is closely intertwined with the feasibility of fine-tuning or re-training. It has been observed that even the random removal of components can ultimately restore the original performance through adequate re-training~\citep{liu2022unreasonable,mittal2019studying}. However, fine-tuning encounters significant challenges when applied to LLMs, rendering pruning less effective. Notably, Wanda~\citep{sun2023simple} achieves remarkable performance by augmenting input activation with weight magnitude, underscoring the critical importance of preserving outlier features in LLM pruning. Considering the vital role that outliers play in the context of LLMs~\citep{kovaleva2021bert,dettmers2022llm} and the success of Wanda, we conjecture that the performance of different pruning methods has a strong correlation with their ability to preserve outlier features. To assess our conjecture, we undertake several preliminary investigations outlined below based on Layerwise Outlier Distribution.

\subsection{Empirical Study}
\label{sec:empirical_study}

\textbf{Layerwise Outlier Distribution \texttt{(LOD)}.}  
\looseness -1 Our preliminary studies are predominantly based on Layerwise Outlier Distribution \texttt{(LOD)}, a concept used to measure the across-layer outlier distribution. Since we focus on weight pruning in this paper, we opt to prioritize the weight outliers instead of the activation outliers, which are identified as weights whose outlier scores are at least $\mathbf{M}$ times larger than the mean. The outlier score of weight is calculated as the accumulation of all input features connected to that weight, multiplied by its magnitude, which also serves as the pruning metric used by Wanda~\citep{sun2023simple}. By measuring the ratio of weight outliers in each layer, we can obtain the  \texttt{LOD} of the whole model.

To formalize our approach, let us consider the input of a layer as $\mathbf{X}$ with dimensions $(N\times L, C_{\texttt{in}})$, where $N$ and $L$ represent the batch and sequence dimensions, respectively; and the weight matrix $\mathbf{W}$ has dimensions $(C_{\texttt{out}}, C_{\texttt{in}})$. The outlier score of weight $\mathbf{W}_{\texttt{ij}}$ is computed as $ \mbf{A}_{\texttt{ij}} =\|\mbf{X}_{\texttt{j}}\|_{2} \cdot |\mathbf{W}_{\texttt{ij}}|$, which is the aggregation of all input features connected to weight $\mathbf{W}_{\texttt{ij}}$, multiplied by its magnitude $|\mathbf{W}_{\texttt{ij}}|$. Here, $\|\mbf{X}_{\texttt{j}}\|_{2}$ is the $\ell_{2}$ norm of input feature connected to the weight. This computation is performed across all $N\times L$ tokens. Subsequently, after obtaining the outlier score for all weights, we proceed to calculate the ``outlier ratio" of $\mbf{A}$ by identifying elements whose magnitude is $\mathbf{M}$ times greater than the averaged value in each layer. We empirically find that {$\mathbf{M}=5$ or $\mathbf{M}=7$ usually works well to sketch the distribution of weight outliers. This process enables us to derive a vector, denoted as \texttt{LOD} = $[D^1, D^2, ... , D^n]$, where $D^l$ characterizes the outlier distribution of layer $l$. That is,
\begin{equation}
    D^l = \frac{\sum_{i=1}^{C_{\texttt{out}}}\sum_{j=1}^{C_{\texttt{in}}}\mathbb{I}(\mbf{A}^l_{\texttt{ij}} > \mbf{M} \cdot \mbf{\bar{A}}^l)}{C_{\texttt{in}}C_{\texttt{out}}}
\end{equation}
where $\mbf{\bar{A}}^l$ is the mean of $\mbf{A}^l$ and $\mathbb{I}(\cdot)$ is the indicator function, returning 1 if $\mbf{A}^l_{\texttt{ij}}$ is larger than $\mbf{M} \cdot \mbf{\bar{A}}^l$, else 0.
Based on \texttt{LOD}, we conduct three empirical studies outlined below to better understand the effect of current LLM pruning approaches on outliers.

\textbf{Empirical Study I: Dense LLMs vs. \texttt{LOD}.} To investigate whether sparsifying LLMs necessitates differential treatment of individual layers, we employ \texttt{LOD} to gauge the layerwise distribution of outliers within dense LLMs. If \texttt{LOD} in dense LLMs exhibits a relatively uniform pattern, it suggests that we do not need a non-uniform layerwise sparsity ratio to preserve outliers, and vice versa. We assess the \texttt{LOD} with LLaMA-7B, 13B, and 30B. 

\begin{table}[h]
\centering
\caption{Effects of various pruning methods on Layerwise Outlier Distribution (\texttt{LOD}) and Perplexity with LLaMA-13B on WikiText. \texttt{LOD} is calculated as the \textit{summation} across all layers with $\mathbf{M}=7$. }
\vspace{-0.8em}
\resizebox{0.45\textwidth}{!}{
\begin{tabular}{l|l|c|c|c}
\toprule
 \textbf{Sparsity} &\textbf{Method} & \texttt{LOD} (\%) $\uparrow$ &  $\Delta$\texttt{LOD} (\%) $\uparrow$ & \textbf{Perplexity} $\downarrow$ \\
\midrule
  & Dense & 5.432 & - & 5.090 \\
\midrule
&Wanda  &  5.716 & 0.284 &  55.900 \\  
70\%  &SparseGPT & \bf 6.645 & \bf 1.213 & \bf 19.235\\  
&Magnitude & 5.322 & -0.110 &  84539.445 \\

\midrule
&Wanda  &  5.433 & 0.001 &   8.761 \\  
60\%  &SparseGPT & \bf 6.044 & \bf 0.612 & \bf 8.458 \\  
&Magnitude & 5.322 & -0.110 &  229.451 \\
\bottomrule
\end{tabular}}
\label{tab:Lod_sparse_metric}
\end{table}

\begin{table}[t]
\centering
\caption{WikiText perplexity with LLaMA-7B of various pruning granularity.}
\vspace{-0.8em}
\resizebox{0.49\textwidth}{!}{
\begin{tabular}{l| c| c| ccccccc}
\toprule
\multirow{2}{*}{\textbf{Method}} & \bf Layerwise & \bf Output  & \multicolumn{7}{c}{\bf Sparsity}  \\

 & \bf Uniform & \bf Balanced  & 10\% & 20\%  &  30\% & 40\% & 50\% & 60\%  & 70\% \\
\midrule
Wanda &  \cmark & \cmark & 5.697 & 5.817 & 5.999 & 6.388 & 7.260 & 10 & 86 \\

Wanda & \cmark & \xmark &  5.695&     5.819&     6.029&     6.572&     7.942&    20&
          238 \\  

Wanda & \xmark & \xmark & 14.117& 3134&10293&10762&14848&17765&  5147 \\

\midrule

Magnitude & \cmark & \cmark  &  5.803 & 6.018 & 6.622 & 8.041 & 13.349 & 152 & 25304 \\

Magnitude & \cmark & \xmark  & 5.806 & 6.020 & 6.669 & 8.601 & {17.287} & 559 & 48419 \\

Magnitude & \xmark & \xmark &  5.821&     6.111&     7.012&     9.825&    48.627& 38335&
        29283 \\

\bottomrule
\end{tabular}
}
\vspace{-2em}
\label{tab:Lod_sparse_gra}
\end{table}

\textbf{Empirical Study II: Pruning Metric vs. \texttt{LOD}.} We further delve into the impact of different pruning metrics on \texttt{LOD}. The primary objective of this study is to explore whether there exists a robust correlation between the performance of various pruning methods and their ability to preserve outliers. To achieve this, we \textit{aggregate} the \texttt{LOD} values across layers for various LLM pruning methods, including magnitude, Wanda, and SparseGPT, and compare them with their dense counterparts. To mitigate the influence of pruning on the mean of outlier score $\mbf{A}$, we use the pre-pruning mean value to measure the outlier score after pruning. Subsequently, the number of outlier weights after pruning is then divided by the total number of weights (including both zero and non-zero weights) to obtain the updated weight outlier ratio.
Doing so helps avoid the impact of pruning on the mean outlier score, ensuring a precise evaluation of alterations in the outlier ratio. All sparse models are pruned with uniform layerwise sparsity. These experiments are conducted using LLaMA-13B at sparsity levels of 60\% and 70\% with $\mathbf{M}=7$.


\looseness=-1 \textbf{Empirical Study III: Pruning Granularity.} It is well-established that non-uniform or global layerwise sparsity often leads to more accurate sparser networks at high sparsity than the uniform layerwise sparsity for pre-LLM pruning. However, endeavors unanimously point out that uniform sparsity is more favorable for LLM pruning. To provide more insights into these two seemingly contradictory arguments, we study the effect of various pruning granularities on LLMs. Specifically, we study two sets of pruning granularities: (1) \textbf{Across different layers,} we compare the performance of uniform pruning and global pruning; (2) \textbf{Within the same layer,} we study the output-imbalanced sparsity used by SparseGPT against the output-balanced sparsity adopted by Wanda. Output-balanced sparsity eliminates the same amount of weights for all outputs. We conduct experiments with magnitude pruning and Wanda using LLaMA-7B. While this part of the study is irrelevant to LOD, we place it here due to the crucial role of pruning granularity.

\looseness=-1 \textbf{Results:} We present our findings from Study 1-3, in Figure~\ref{fig:layerwise}, Table~\ref{tab:Lod_sparse_metric}, and Table~\ref{tab:Lod_sparse_gra}, respectively. These results provide
positive support for our conjecture, and we summarize the key observations below:

\textbf{\circled{1} \texttt{LOD} of dense LLMs exhibits a highly non-uniform distribution across layers. } In essence, the distribution of dense LLMs shown in Figure~\ref{fig:layerwise} loosely follows a ``U" shape, with notable proportions at both ends, while the central region displays a monotonic descending trend.
This finding validates our conjecture that individual layers need unique consideration during the pruning procedure. Employing uniform pruning across all layers would inevitably disrupt the outlier structure in layers characterized by a large outlier ratio, such as those layers  at the beginning or end of models.

\textbf{\circled{2} The performance of sparse pruning methods on LLMs is closely correlated with their ability to retain outlier features.} Leading pruning techniques like Wanda and SparseGPT all excel in outlier,  resulting in an overall increase in \texttt{LOD}. In contrast, the naive baseline of magnitude pruning performs no better than random selection at 70\% sparsity, as evidenced by a negative change of -0.110 in \texttt{LOD}, indicating the removal of important outliers. It is interesting to see that despite SparseGPT not being explicitly designed for outlier preservation, it achieves the highest \texttt{LOD} as well as performance, providing further insight into the underlying reason for its success. 
The potential explanation could be that the weight update involved within SparseGPT contributes to the increase in \texttt{LOD}.

\textbf{\circled{3} Pruning with coarser granularity results in diminished performance.} In general, we observe a consistent trend of improved perplexity as the pruning granularity becomes finer, transitioning from global layerwise sparsity to uniform layerwise sparsity at the macro level, and from output imbalanced sparsity to output balanced sparsity at the micro level. These findings align with the conclusions presented by~\citet{sun2023simple}. This observation suggests the importance of a nuanced design of pruning ratios to mitigate aggressive sparsity differences among different components in LLMs. This motivation led us to constrain the sparsity ratio of each layer to fluctuate only around the target sparsity, with a hyperparameter $\lambda$ introduced in Section \ref{sec:owl}.




\begin{figure}[t]
\vspace{-0.5em}
\centering

    \subfigure{
        \includegraphics[width=0.35\textwidth]{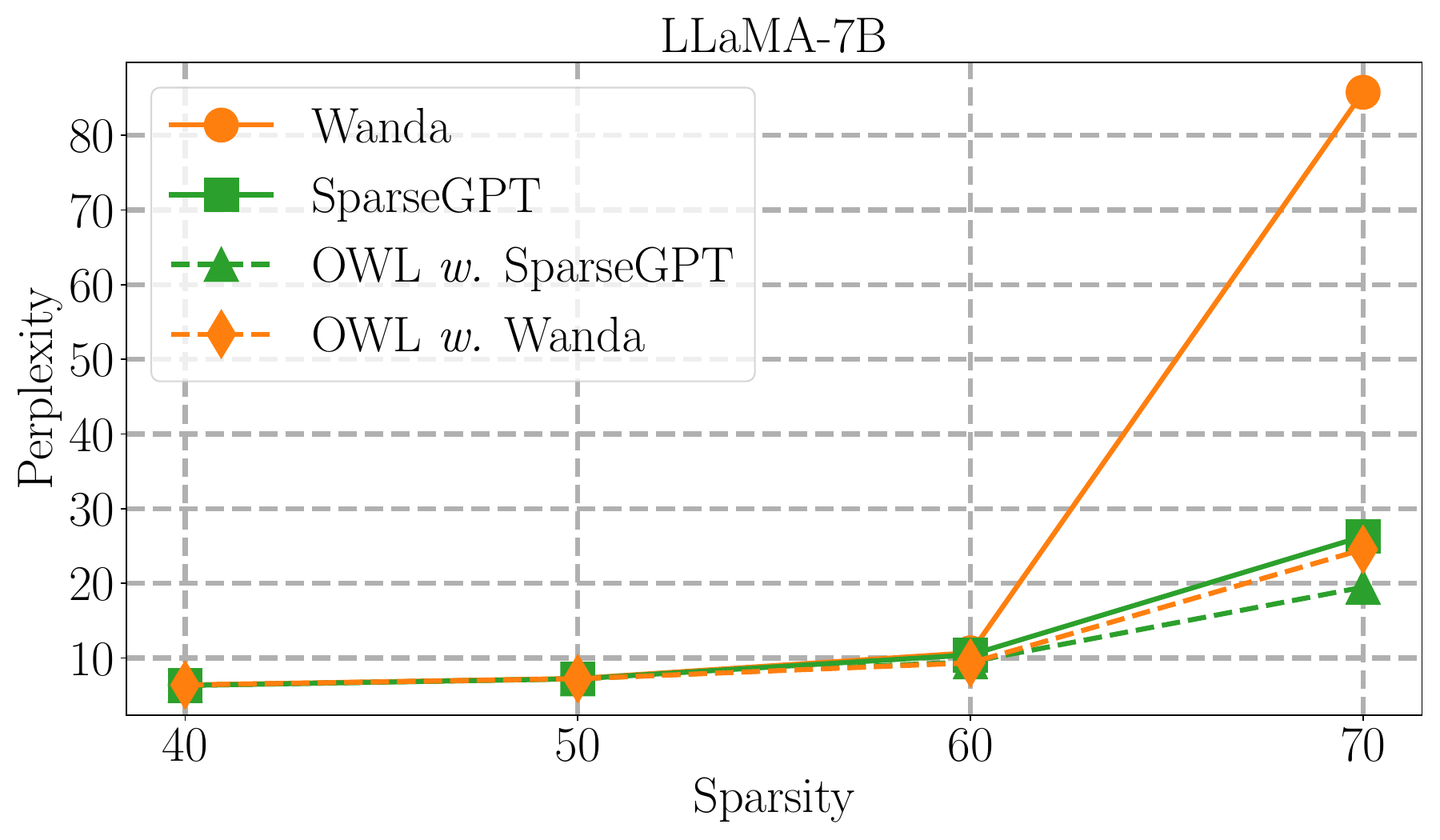}
    }
    \vspace{-1em}
    \subfigure {
        \includegraphics[width=0.35\textwidth]{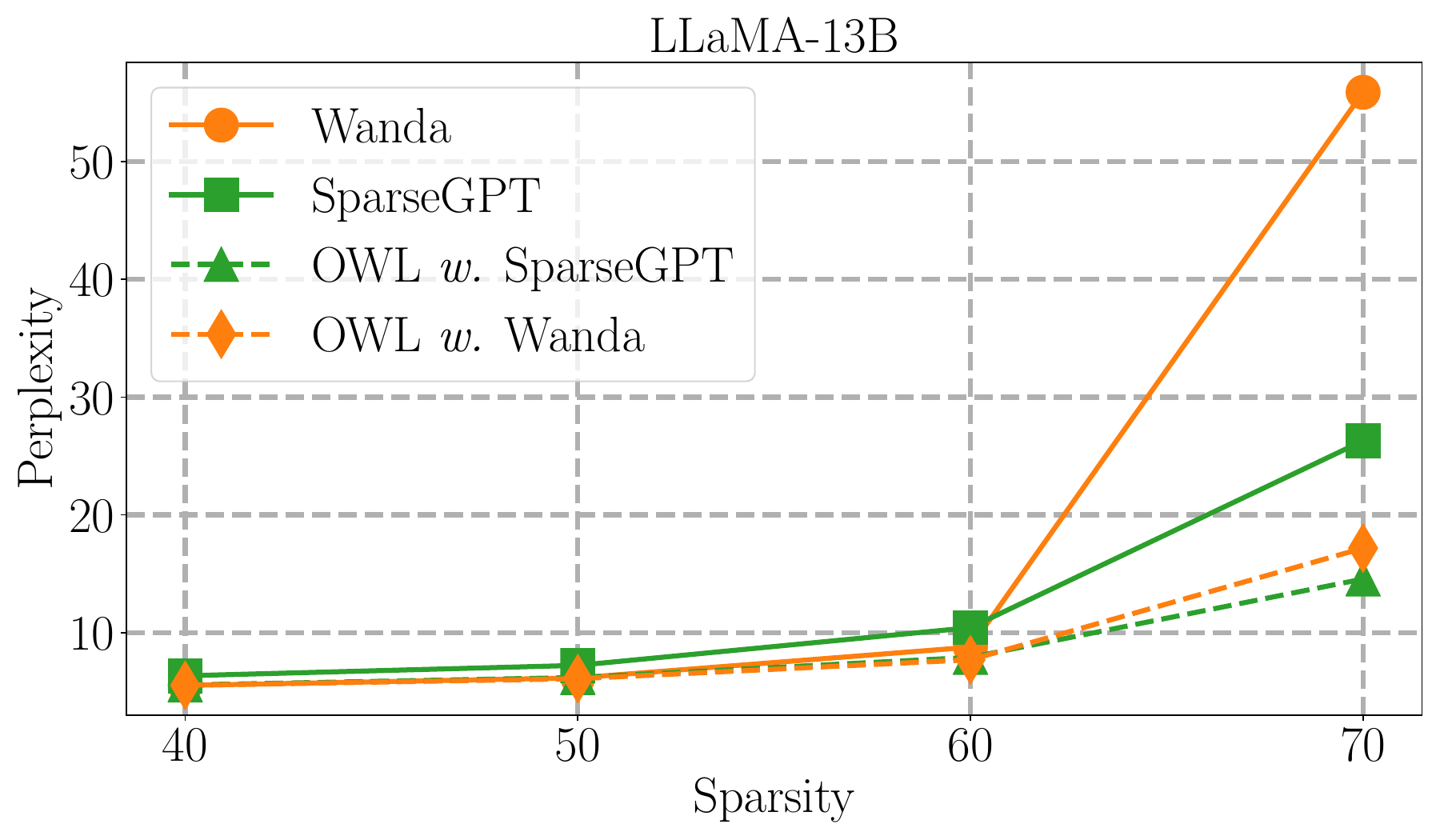}
    }

\vspace{-1em}
\caption{WikiText validation perplexity of LLaMA-7B and LLaMA-13B pruned by various approaches.}
\vspace{-2em}
\label{fig:LLaMA_7B}
\end{figure}









\begin{table*}[h]
\centering
\vspace{-0.1cm}
\caption{WikiText validation perplexity of pruning methods for LLaMA-V1 family and OPT-6.7B at 70\% sparsity. The best performance method is indicated in \textbf{bold}, and the gain in perplexity achieved by OWL is highlighted in blue.}
\vspace{-0.8em}
\resizebox{0.8\textwidth}{!}{
\begin{tabular}{l|c|c|llll | l}
\toprule 
\bf Method & \bf Layerwise  & \bf  Weight & \multicolumn{4}{c|}{\bf LLaMA-V1} & \bf OPT \\
    & \bf Sparsity & \bf  Update & \bf 7B & \bf  13B & \bf 30B & \bf 65B  & \bf 6.7B \\
    \midrule
    Dense & - &  -  &  5.68 & 5.09 & 4.10 & 4.77  &   10.13\\
    \midrule
    Magnitude  & Uniform & \xmark & 48419.12  & 84539.45 & 977.73 &  46.89 &  290985.03  \\
    \midrule
    Wanda & Uniform & \xmark  & 85.77 & 55.90 &   17.37  &   15.23 &  162.92 \\
    OWL \textit{w.} Wanda  & Non-Uni & \xmark  & \bf 24.55 \textcolor{blue}{\scriptsize (-61.22)} & \bf  17.17 \textcolor{blue}{\scriptsize (-38.73)} & \bf  10.75 \textcolor{blue}{\scriptsize (-6.62)} & \bf 8.61 \textcolor{blue}{\scriptsize (-6.62)} & \bf  40.22 \textcolor{blue}{\scriptsize (-120.70)}\\
    \midrule
    SparseGPT & Uniform & \cmark  &26.30   & 19.24 &  12.56 & 10.45 &  \bf 20.29 \\
    OWL \textit{w.} SparseGPT   & Non-Uni & \cmark  & \bf 19.49 \textcolor{blue}{\scriptsize (-6.81)} & \bf 14.55 \textcolor{blue}{\scriptsize (-4.69)}  & \bf 10.28 \textcolor{blue}{\scriptsize (-2.28)}  & \bf 8.28 \textcolor{blue}{\scriptsize (-0.64)} &   22.48 \textcolor{blue}{\scriptsize (2.19)}\\ 
     \bottomrule
\end{tabular}}
\label{tab:LM_unstr_sparsity}
\vspace{-0.7em}
\end{table*}

\subsection{Outlier Weighed Layerwise Sparsity (OWL)} 
\label{sec:owl}
The above empirical studies underscore the critical significance of preserving outliers in the context of LLM pruning. Consequently, it becomes imperative to implement layerwise pruning strategies that take into account the non-uniform distribution of outliers across different layers. However, global pruning can be costly and lead to the collapse of outliers, resulting in significant performance degradation. On the other hand, uniform pruning does not adequately consider the highly non-uniform distribution of outlier features across various layers. This negligence inevitably disrupts the structure of outliers in layers characterized by a substantial outlier ratio, particularly at high sparsity levels. Therefore, there is a need of an ideal layerwise sparsity that aligns effectively with the layerwise outlier distribution while maintaining computational and memory efficiency.

To address this issue, we propose a novel layerwise sparsity ratio strategy, referred to as  \textbf{O}utlier-\textbf{W}eighed \textbf{L}ayer-wise sparsity (\textbf{OWL}) explicitly tailored for LLMs, which can better coordinate with the outlier distribution by taking the layerwise outlier ratio into consideration. Given a $l$-layer large language model with a target model sparsity $S$, we aim to calculate the target layerwise sparsity $[S^1, S^2, ... , S^n]$.  We first calculate \texttt{LOD}, $\ \mbf{D} = [D^1, D^2, ... , D^n]$, based on the approach proposed in Section~\ref{sec:empirical_study}. Guided by the principle that layers with a higher proportion of outliers should have a lower sparsity, we set $ S_i \propto 1-D_i $. Additionally,  we introduce a hyperparameter $\lambda$ which constrains the layerwise sparsity to fall within a small range, specifically, $S_i \in [S-\lambda, S+\lambda]$, while maintaining an average sparsity of $S$ across all layers. This helps prevent excessive differences in sparsity between layers, ensuring a robust performance. This constraint is inspired by the insights gained from ``Empirical Study III'' which highlight the detrimental impact of overly aggressive layerwise sparsity, akin to global pruning, on sparse LLMs. To obtain a favorable number for $\lambda$ and $M$, we conduct a small hyperparameter sweep within the range of $\lambda \in$ [0.02, 0.05, 0.08, 0.1, 0.2] and for $M \in$ [3, 5, 7, 10]. Note that we assign a distinct pruning ratio for each Transformer block instead of each layer, as the former provides better performance (see Appendix \ref{app:per-block} for more detail).
The visualization of our layerwise sparsity ratio is demonstrated in  Figure~\ref{fig:layerwise}, where we can see that the layerwise sparsity level of OWL nuancedly aligns well with the model's \texttt{LOD}. 

\begin{table*}[h]
  \centering
    \caption{
  Accuracies (\%) for 7 zero-shot tasks with 70\% sparsity using LLaMA-V1 family.
  }
\vspace{-0.8em}
\resizebox{0.8\textwidth}{!}{
  \setlength{\tabcolsep}{5.5pt}
  \begin{tabular}{ccccccccccc}
  \toprule
\bf Params  & \hspace{-0.4cm} \bf Method & \hspace{-0.2cm} \bf BoolQ & \bf RTE & \hspace{-0.35cm} \bf HellaSwag  & \hspace{-0.35cm}\bf  WinoGrande & \hspace{-0.35cm} \bf ARC-e & \bf ARC-c & \bf OBQA & \bf Mean \\
  \midrule
  \multirow{6}{*}{7B}   & Dense & $75.14$ & $66.43$ & $74.80$ &$70.01$ & $67.67$ & $41.38$ & $41.40$ & $62.40$ \\
  \cmidrule{2-10}
  & Magnitude   & $38.29$ & $52.71$ & $24.68$ & $51.46$ & $26.98$ & $22.35$ & $25.80$ & $34.61$  \\ 

  & Wanda   & $55.11$ & $57.40$ & $31.83$ & $51.38$ & $34.22$ & $19.80$ & $26.00$  & $39.39$ \\ 
    \gr &  OWL \textit{w.} Wanda & \bf 62.48 & \bf 58.48 & \bf 44.79 & \bf 58.72 & \bf 45.03 & \bf 26.19 & \bf 29.60 & \bf 46.47\\ 
    \cmidrule{2-10}
      & SparseGPT  & 64.53  & \bf  53.79  & 42.11  & $58.64$  & 43.06 & 24.57 & 27.80 & 44.93 \\ 
    \gr & OWL \textit{w.} SparseGPT & \bf 67.13 &  53.43 & \bf 48.56& \bf 62.03 & \bf 45.41& \bf 27.65 & \bf 32.00& \bf 48.03 \\
  \midrule

  \multirow{6}{*}{13B} & Dense & $77.86$ & $70.40$ & $78.08$ & $72.77$ & $69.19$ & $47.18$ & $43.80$ & $65.61$ \\
  \cmidrule{2-10}
  & Magnitude  & $52.94$ & $50.54$ & $27.67$ & $50.91$ & $28.24$ & $23.38$ & $24.80$ & $36.93$ \\

  & Wanda & $61.71$ &  \bf 52.71 & $34.31$ & $52.33$ & $37.16$ & $20.90$ & $29.60$ & $41.25$ \\
    \gr & OWL \textit{w.} Wanda & \bf 62.69 &  \bf 52.71 & \bf 51.03 & \bf 63.14  & \bf 49.54 & \bf 28.67 & \bf 34.40 & \bf 48.88 \\
    \cmidrule{2-10}
      & SparseGPT  & \bf 66.94 & 52.71 & 47.91  & 62.90 & 45.03 & 27.99 & 35.20 & 48.38 \\
    \gr & OWL \textit{w.} SparseGPT & 64.95 & \bf 53.07 & \bf 54.39& \bf 66.54 & \bf 48.86& \bf 30.12& \bf 38.00& \bf 50.85 \\
  \midrule

  \multirow{6}{*}{30B} & Dense & 82.69 & 66.79 & 81.19 & 75.85 & 73.48 & 50.77 & 44.60 & 67.91 \\
  \cmidrule{2-10}
  & Magnitude  & 39.14 & 46.21 & 24.31 & 52.33 &  24.66 & 22.87 & 29.00 & 34.07 \\

  & Wanda & $66.12$ & \bf 57.76 & $58.84$ & $67.32$ & $59.26$ & $33.11$ & \bf 40.20 & $54.66$\\
    \gr & OWL \textit{w.} Wanda & \bf 66.42  & 52.35 & \bf 62.94  & \bf 69.30  & \bf 61.83 & \bf 35.84 & 40.00 & \bf 55.53 \\
      \cmidrule{2-10}
      & SparseGPT  & 66.51 & \bf 63.90 & 60.38 & 69.85 & 58.54 & 33.70 & 40.60 & 55.78 \\
    \gr & OWL \textit{w.} SparseGPT & \bf 67.58 & 58.48 & \bf 64.88 & \bf 70.72 & \bf 60.82 & \bf 35.07 & \bf 42.20 & \bf 57.11 \\
  \midrule 

  \multirow{6}{*}{65B} & Dense & 84.86 & 69.68 & 82.94 & 77.35 & 75.08 & 52.56 & 44.20 & 69.52 \\
  \cmidrule{2-10}
  & Magnitude  & 52.17 & 54.87 & 49.87 & 56.67 & 49.71 & 30.63 & 38.80 & 47.53 \\

  & Wanda  & 76.30 & 56.68 & 61.26 & 70.48 & 63.47 & 35.67 & 39.40 & 57.61 \\
    \gr & OWL \textit{w.} Wanda  & \bf 80.12 & \bf 58.84 & \bf 66.16 & \bf 73.56 & \bf 65.45 & \bf 39.93 &\bf  42.20 & \bf 60.89 \\
      \cmidrule{2-10}
      & SparseGPT   & 80.64 & 59.57 & 66.42 & 72.61 &\bf 60.52 & 38.57 & \bf 40.80 & 59.88 \\
    \gr & OWL \textit{w.} SparseGPT  & \bf 82.63 & \bf 67.15 & \bf 68.52 & \bf 75.06 & 60.10 & \bf 39.59 & 39.00 & \bf 61.72 \\
  \midrule 
  \end{tabular}
  }
 \vspace{-0.5em}
  \label{tab:zero_shot_main}
\end{table*}



\begin{table}[h]
\centering
\vspace{-0.5em}
\caption{WikiText perplexity of ``OWL \textit{w.} SparseGPT'' with LoRA fine-tuning.}
\vspace{-0.8em}
\resizebox{0.33\textwidth}{!}{
\begin{tabular}{lccc}
\toprule
\bf Method &\bf Model  &\bf Sparsity  &\bf Perplexity  \\ %
\midrule
 Without FT  &  7B  & 0.7  & 19.49 \\
 With FT &  7B  &  0.7 & 11.15 \\
\midrule
 Without FT  &  13B  & 0.7  & 14.55 \\
 With FT &  13B  &  0.7 & 9.0 \\
\bottomrule
\end{tabular}}
\label{tab:FT}
\end{table}

\begin{table}[h]
\centering
\vspace{-0.8em}
\caption{Comparison of time overhead used for computing the pruning metric across layers of LLaMA (in seconds).}
\vspace{-0.8em}
\resizebox{0.33\textwidth}{!}{
\begin{tabular}{lccccc}
    \toprule
    \bf Method &  \bf 7B  & \bf 13B  &  \bf 30B &  \bf 65B \\ 
    \midrule
    SparseGPT   & 266 & 436 & 869 & 1395 \\
   \gr  OWL \textit{w.} SparseGPT   & 268 & 438 & 870 & 1397 \\
    Wanda & 0.3 & 0.4  & 1.1  & 1.8\\
     \gr  OWL \textit{w.} Wanda   & 0.3 & 0.5 & 1.2 & 2.0 \\
    \hline
    \end{tabular}
}
\vspace{-1em}
\label{tab:time_overhead}
\end{table}

\vspace{-0.5em}
\section{Experiments}
\subsection{Main Experiments}

\textbf{Models and Dataset.} We assess OWL's performance across a range of LLMs, encompassing the LLaMA-V1 model family~\citep{touvron2023llama2} with parameter counts ranging from 7 billion to 65 billion, as well as OPT-6.7B~\citep{zhang2022opt}. Our evaluation protocol aligns with established LLM pruning methodologies~\citep{frantar2023massive,sun2023simple}, encompassing assessments of language modeling proficiency and zero-shot capabilities of sparse LLMs. Specifically, we measure the Perplexity metric on the WikiText~\citep{wikitext103} validation dataset for language modeling performance, and employ the Accuracy metric for zero-shot evaluations on seven common sense benchmarks, including BoolQ~\citep{clark2019boolq}, RTE~\citep{glue}, HellaSwag~\citep{hellaswag}, WinoGrande~\citep{sakaguchi2019winogrande}, ARC Easy and Challenge~\citep{clark2018think}, and OpenbookQA~\citep{mihaylov2018can}.

\textbf{Baselines.} We choose the three current LLM-pruning baselines, including magnitude~\citep{jaiswal2023emergence}, SparseGPT~\citep{frantar2023massive}, Wanda~\citep{sun2023simple}. Magnitude pruning serves as a naive baseline for LLMs, with an expected sharp decline in performance at modest sparsity levels, typically ranging from 10\% to 30\%. SparseGPT and Wanda, on the other hand, are established baselines known for their ability to maintain reasonable performance even at relatively high sparsity levels, typically around 50\% to 60\%. Notably, in contrast to our approach, all baseline methods employ with uniform layerwise sparsity. We primarily focus on high sparsity levels, not falling below 50\%, as regions with low sparsity pose challenges for existing sparse GPU kernels to outperform their dense counterparts~\citep{gale2020sparse}. To ensure equitable comparisons, we have employed the identical set of calibration data as utilized by SparseGPT and Wanda for model pruning, \emph{i.e.,} comprising 128 sequences with 2048 tokens for each, randomly sampled from the first shard of the C4~\citep{raffel2020exploring} dataset. We incorporate OWL directly into Wanda and SparseGPT, resulting in two variants: ``OWL \textit{w.} Wanda'' and ``OWL \textit{w.} SparseGPT''. The only distinction between these variants lies in their layerwise sparsity ratios, with OWL providing a more tailored layerwise sparsity in this regard. Hyperparameters are shared in Appendix~\ref{app:hyper}.

\textbf{Language Modelling.} We first report the performance of various LLM pruning methods on language modelling with WikiText. The results is presented in Table~\ref{tab:LM_unstr_sparsity} and Figure~\ref{fig:LLaMA_7B}. We summarize the key observation below:

\textbf{\circled{1} OWL serves as a general layerwise sparsity method suitable for various scenarios.} As illustrated in Table~\ref{tab:LM_unstr_sparsity}, OWL exhibits effectiveness across different pruning methods (such as Wanda and SparseGPT), architectural variants (including LLaMA-V1 and OPT), and diverse model sizes (ranging from 7B, 13B, 30B, to 65B parameters), resulting in substantial reductions in perplexity scores. Notably, even when applied to SparseGPT, a strong pruning method incorporating second-order information, OWL still achieves significant perplexity reductions, exemplified by a reduction of 6.81 for LLaMA-7B.
 
\textbf{\circled{2} The benefits of OWL  increase significantly as model size decreases.} There is a clear trend that the performance gain of OWL monotonically increases as LLaMA-V1 scales down from 65B to 7B. While the performance improvement of OWL \textit{.w} Wanda for LLaMA-65B is already encouraging i.e., 6.62, it achieves a remarkable gain of 61.22 for LLaMA-7B, resulting in a reasonable 24.55 perplexity.





\textbf{Zero-Shot Tasks.} 
While perplexity is a widely used metric for language modeling, it primarily serves as a statistical measure of how confidently a language model predicts a text sample and does not necessarily align with the quality of the generated text \citep{jaiswal2023compressing}. To draw more robust conclusions, we conducted experiments to evaluate the zero-shot ability of various sparse LLMs on diverse zero-shot downstream tasks with prompting. These experiments were performed using the LLaMA-V1 family at 70\% sparsity, and the results are presented in Table~\ref{tab:zero_shot_main}. It's noteworthy that OWL consistently improves accuracy across nearly all settings, with very few exceptions on RTE dataset. For example, OWL achieves an average perplexity gain of 4.72 and 2.19 over 7 tasks and 4 model sizes compared to Wanda and SparseGPT alone, respectively. This result highlights the promise of OWL still holds for more challenging tasks.


\begin{table*}[h]
\centering
\caption{End-to-end decode latency speedup of LLaMA-V2-7B-chat-hf using OWL with the DeepSparse \citep{NM} inference engine.}
\vspace{-0.8em}
\resizebox{0.8\textwidth}{!}{
\begin{tabular}{@{}lcccccccccccc@{}}
\toprule
\bf Sparsity & \bf Dense & \bf 10\% &  \bf 20\% &  \bf 30\% & \bf 40\% & \bf 50\% & \bf 60\% & \bf 70\% & \bf 80\% & \bf 90\% \\
\midrule
Latency (ms) & 213.83 & 216.86 & 221.62 & 218.01 & 167.54 & 121.25& 
 101.41 & 81.89 & 64.57 & 54.24 \\
Throughput (tokens/sec) & 4.68 & 4.61 & 4.51 & 4.59 & 5.97 & 8.25	& 9.86 & 12.21 & 15.48 & 18.43  \\ 
Speedup & 1.0x& 1.0x& 1.0x& 1.0x& 1.3x&  1.8x&  2.1x&  2.6x&  3.3x& 
 3.9x \\ 
\bottomrule
\label{tab:speedup}
\end{tabular}}
\vspace{-1em}
\end{table*}

\begin{table*}[t]
\centering
\caption{WikiText validation perplexity of LLaMA-7B with various layerwise sparsity using Wanda.}
\vspace{-0.8em}
\resizebox{0.8\textwidth}{!}{
\begin{tabular}{l*{8}{c}}
\toprule
\bf Sparsity/Perplexity &  \bf 10\% & \bf 20\% & \bf 30\% & \bf 40\% & \bf 50\% & \bf 60\% & \bf 70\% & \bf 80\% \\
\midrule
Global & 14.11& 3134&10293&10762&14848&17765&  5147 &39918.56 \\
ER-plus & 5.70 & 5.82 & 6.05 & 6.62 & 8.00 & 14.04 & 229.17 & 6013.91 \\
ER & 5.69 & 5.80 & 6.02 & 6.55 & 7.74 & 12.16 & 112.03 & 11151.18 \\
Uniform & 5.69 & 5.81 & 5.99 & 6.38 & 7.26 & 10.70 & 85.77 & 3499.88 \\
OWL-inverse & 5.72 & 5.83 & 6.04 & 6.51 & 8.03 & 26.05 & 822.23 & 9616.08 \\
 \gr OWL (ours)  & 5.70 & 5.80 & 6.01 & 6.39 & 7.22 & 9.35 & 24.54 & 1002.87 \\
\bottomrule
\end{tabular}
}\vspace{-0.8em}
\label{tab:com_common_sparsity}
\end{table*}

\textbf{Fine-tuning Performance.} 
We also explore the impact of fine-tuning on the performance recovery of OWL. In alignment with Wanda, we utilize LoRA~\citep{hu2021lora} as our fine-tuning method and refrain from merging the adapter back to preserve the sparse pattern. We fine-tune models pruned by ``OWL + SparseGPT'' using only a minimal 30,000 tokens from the C4 training dataset. Remarkably, our results demonstrate that the perplexity drop caused by aggressive pruning can be significantly narrowed through a very short time of fine-tuning, reducing the perplexity from 19.49 to  11.15 of LLaMA-7B and from 14.55 to 9.0 for LLaMA-13B. We anticipate achieving a significantly lower perplexity by employing more advanced sparse fine-tuning approaches \citep{zimmer2023perp}, or by extending the fine-tuning duration.

\textbf{Pruning Efficiency.} Since we utilize the pruning metric of Wanda to determine our layerwise sparsity, the computational complexity of OWL is comparable to that of Wanda. To demonstrate this, we measure the total pruning time, excluding the forward pass process, following the methodology outlined by~\citet{sun2023simple}. These results were obtained using NVIDIA A100 GPUs.
Our results in Table~\ref{tab:time_overhead} indicate that OWL incurs negligible time
overhead (maximum 2 seconds) relative to previous pruning approaches.

\looseness -1 \textbf{Inference Speedup.} We analyze the speedup achieved by OWL, as presented in Table \ref{tab:speedup}. The reported speedups correspond to end-to-end decode latency using LLaMA-V2-7B-chat-hf 
in the DeepSparse inference engine \cite{NM} on an Intel Xeon Platinum 8360Y CPU with 36 cores. It is evident that OWL delivers a significant inference speedup compared to the dense model, reaching 2.6$\times$ at 70\% sparsity. Notably, the speedup gain becomes even more substantial with higher sparsity e.g., around 4$\times$ at 90\% sparsity, showcasing additional motivation for future endeavors targeting extreme sparsity.

\subsection{More Advanced LLMs}
\begin{table}[h]
\centering
\caption{WikiText Perplexity of Various LLMs.}
\vspace{-0.5em}
\resizebox{0.45\textwidth}{!}{
\begin{tabular}{l|c|cccc}
    \toprule
    & & \multicolumn{3}{c}{\bf Sparsity} \\
    \bf Method &  \bf Models  & \bf 60\%  &  \bf 70\% &  \bf 80\% \\ 
    \midrule
    Wanda   & LLaMA-V2-7B & 11.63 & 57.10 & 3221.74 \\
   \gr  OWL \textit{w.} Wanda     & LLaMA-V2-7B & \bf 10.30 & \bf 29.07 & \bf 523.30 \\
    
    Wanda & Vicuna-7B & 11.93 & 57.98 & 2347.49 \\
    \gr  OWL \textit{w.} Wanda   & Vicuna-7B & \bf 10.58 & \bf 34.44  & \bf 719.11  \\
    Wanda   & Mistral-7B & 11.27 & 60.67 & 306.26 \\
   \gr  OWL \textit{w.} Wanda     & Mistral-7B & \bf 10.29 & \bf 41.02 & \bf 220.89 \\
    \bottomrule
    \end{tabular}
}
\label{tab:more_advanced_llms}
\end{table}
To examine if the effectiveness of OWL is robust across various LLMs, we also apply OWL to more advanced LLMs including LLaMA-V2-7B-chat-hf \cite{touvron2023llama2}, Vicuna-7B, and Mistral-7B \cite{jiang2023mistral}. Table \ref{tab:more_advanced_llms} further illustrates the efficacy and robustness of OWL as a layerwise sparsity technique across a spectrum of Large Language Models (LLMs). Moreover, Mistral appears to preserve the perplexity better at high sparsity such as 80\%.

\subsection{More Practical Applications of OWL}

Although GPUs provide limited support for unstructured sparsity,  we highlight that OWL not only can be used to determine the sparsity ratio for unstructured pruning but also serves as a general approach to identify layer importance for LLMs, which exhibits substantial potential in many important scenarios. This section provides some preliminary results for low-rank approximation of LLMs. You can find more preliminary results across three more practical scenarios including N:M sparsity, structured pruning, and mixed-precision quantization in Appendix \ref{app:practical_owl}. 

\textbf{SVD Low-rank Approximation.} We expanded the OWL approach to incorporate low-rank compression as follows. In alignment with the OWL methodology, we evaluate the significance of each layer through the LoD. Subsequently, we utilize these LoD scores to assign SVD compression ratios to each layer. A higher LoD score signifies a layer of greater importance, warranting reduced compression to maintain its integrity. For any linear layer $L$ equipped with a weight matrix $W \in \mathbb{R}^{d_1 \times d_2}$ and a designated preserved rank $r$, the compression ratio is defined by $\frac{r}{d_{\text{min}}}$, where $d_{\text{min}}$ is the lesser of $d_1$ and $d_2$. In our evaluation, we present the perplexity metrics on the LLaMA-V1 7B model without any fine-tuning. The results clearly demonstrate that OWL-SVD surpasses uniform SVD low-rank compression across various rank reduction levels.


\begin{table}[h]
\centering
\caption{WikiText Perplexity of LLaMA-V1-7B Model.}
\vspace{-0.8em}
\resizebox{0.45\textwidth}{!}{
\begin{tabular}{lcccccc}
\toprule
\bf Rank Reduction Ratio & \bf 0\% & \bf 20\% & \bf 30\% & \bf 40\% & \bf 50\% & \bf 60\% \\
\midrule
Uniform - SVD & 5.68 & 8.48 & 17.23 & 1909.34 & 13627.03 & 34354.9 \\
 \gr OWL - SVD & 5.68 & \bf 8.25 & \bf 12.92 & \bf 43.02 & \bf 10707.41 & \bf 20118.7 \\
\bottomrule
\end{tabular}}
\label{tab:SVD}
\end{table}

\vspace{-0.4em}
\section{Analysis}
\vspace{-0.3em}
\label{sec:com_lw_sparsity}

\textbf{Comparisons Among Various Layerwise Sparsity.}
\label{sec:com_ls}
We compare OWL layerwise sparsity with multiple commonly used layerwise sparsity, including \textbf{Global}: A global threshold is uniformly applied to all layers to satisfy the overall sparsity requirement, and the specific layerwise sparsity is automatically adjusted based on this threshold. \textbf{Uniform}~\citep{zhu2017prune}: Every layer is pruned with the same target sparsity. \textbf{Erd\H{o}s-R\'{e}nyi (ER)}~\citep{mocanu2018scalable}: The sparsity of the convolutional layer is scaled proportionally to $1 - \frac{n^{l-1}+n^{l}}{n^{l-1}\times n^{l}}$ where $n^l$ refers to the number of neurons/channels in layer $l$.
\textbf{ER-plus}~\citep{liu2022unreasonable}: ER-plus modifies ER by forcing the last layer as dense if it is not while keeping the overall parameter count the same. \textbf{OWL-inverse}: OWL-inverse metric is the inverse variant of OWL, whose outlier ratio is $1 - \texttt{LOD}$.

\looseness=-1 For this study, we apply Wanda to the LLaMA-7B model. The results are presented in Table~\ref{tab:com_common_sparsity}. It is noteworthy that all approaches, except for the Global method, perform satisfactorily when the sparsity level is at or below 40\%. This observation suggests that the region of low sparsity does not provide significant distinctions for performance comparison. However, as the sparsity level exceeds 50\%, discrepancies between the various approaches become evident. Notably, the Uniform and OWL methods emerge as the top-performing approaches, with OWL consistently outperforming the former across all sparsity levels. On the other hand, the ER family of methods appears to be less suitable for LLM pruning. It's worth mentioning that the performance of OWL experiences a significant decline when we invert its outlier ratio, underscoring the effectiveness of \texttt{LOD} in identifying critical layers.

\textbf{Vision Models.} Moreover, we also evaluated OWL in vision models with vision models in Appendix \ref{app:vision}. The performance improvement of OWL on vision models is not as pronounced as observed in LLMs, likely as outliers are not particularly evident in vision models.

\vspace{-0.5em}
\section{Conclusion}
In this paper, we focus on a crucial aspect of LLM pruning that has been overlooked by previous works -- layerwise sparsity ratios. Drawing inspiration from the emergence of outliers, characterized by features exhibiting significantly greater magnitudes compared to others, we introduced a novel layerwise sparsity ratio, Outlier Weighed Layerwise sparsity (OWL). OWL aligns the sparsity ratio with the outlier ratio of each layer to preserve outliers. Notably, our approach demonstrates substantial performance gains, surpassing the state-of-the-art Wanda and SparseGPT by 61.22 and 6.80 perplexity points at 70\% sparsity, respectively. This work represents the first effort to make it applicable in LLM pruning, opening up new avenues for the development of specialized sparse algorithms that can further optimize the deployment of LLMs in practical applications.

\section{Impact Statements}
This paper focuses on pruning large language models (LLMs). Through the implementation of effective layerwise sparsity, we can achieve a reduction of up to 70\% in the parameters of SOTA LLMs while retaining their essential functionality. Consequently, this advancement facilitates the deployment of LLMs on resource-constrained devices, expedites inference processes, and contributes to the sustainability of AI technologies. The significant real speed-up demonstrated in our study underscores the considerable advantages of sparsity, extending beyond GPU to other commodity hardware like CPU and FPGA. This encourages researchers to broaden their focus beyond GPU-centric perspectives, fostering a deeper exploration of the benefits of sparsity across diverse hardware platforms. Moreover, our paper aligns with the imperative of developing energy-efficient and accessible artificial intelligence systems, catering to a wide array of applications and users.

\section{Acknowledgement}

S. Liu is funded by the Royal Society with the Newton International Fellowship. 
Part of this work used the Dutch national e-infrastructure with the support of the SURF Cooperative using grant no. NWO-2023.027/L1 and EINF-2943/L1. Z. Wang is in part supported by the NSF AI Institute for Foundations of Machine Learning (IFML).  We extend our gratitude to Eldar Kurtic for his invaluable effort with the real speedup measurement of the DeepSparse engine.

\nocite{langley00}

\bibliography{example_paper}
\bibliographystyle{icml2024}

\newpage
\appendix
\onecolumn

\section{Vision Model Pruning}
\label{app:vision}

In this appendix, we study if the promise of OWL also holds for vision models. We apply OWL to two commonly used modern vision models, i.e., ConvNeXt-Base \citep{liu2022convnet} and DeiT-Base \citep{touvron2021training} and evaluate them on the ImageNet-1K dataset \citep{deng2009imagenet}. We adopt Wanda as the pruning approach and compare OWL with uniform layerwise sparsity. All models are pruned in one shot without fine-tuning.

\begin{table*}[h]
\centering
\caption{Top-1 accuracy of sparse vision models on ImageNet-1K.}
\label{tab:vision}
\vspace{-0.8em}
\resizebox{0.6\textwidth}{!}{
\begin{tabular}{lccccc}
\toprule
\bf \multirow{2}{*}{Method} & \bf \multirow{2}{*}{Model} & \multicolumn{4}{c}{\bf Sparsity} \\
   & & \bf 50\% & \bf 60\% & \bf 70\% & \bf 80\% \\
\midrule
Wanda & ConvNeXt-Base & 82.72 & \bf 80.55 & 68.18 & \bf 6.44 \\
\gr OWL \textit{w.} Wanda & ConvNeXt-Base & \bf 82.76 & 80.53 & \bf 68.28 & 6.32 \\
\midrule
 Wanda & DeiT-Base & 78.23 & 71.14 & 49.20 & 6.86 \\
\gr  OWL \textit{w.} Wanda & DeiT-Base & \bf 78.40 & \bf 71.76 & \bf 54.24 & \bf 7.98 \\
\bottomrule
\end{tabular}}
\end{table*}

Our findings in Table \ref{tab:vision} reveal that OWL enhances the accuracy of sparse DeiT models in contrast to Wanda. However, for ConvNeXt models, it seems that OWL does not necessarily bring benefits to ConvNeXt (neither increases nor degrades accuracy). Overall, it seems that the performance improvement of OWL on vision models is not as pronounced as observed LLMs. Our hypothesis here is that the phenomenon of outliers is not particularly evident in vision models. According to \citet{puccetti2022outliers}, outliers in LLMs are causally related to high-frequency tokens in pre-training data. These high-frequency tokens are more prevalent in textual datasets but are relatively scarce and challenging to identify within vision datasets. Hence, the phenomenon of outliers, crucial in OWL's effectiveness, may not be as evidently present or impactful within the domain of vision models, contributing to the differing performance improvements between LLMs and vision models.

\section{More Practical Applications of OWL}
\label{app:practical_owl}

OWL serves as a general approach to identify layer importance for LLMs, which exhibits substantial potential in many hardware-friendly scenarios. To examine this, we explore OWL in three hardware-friendly regimes, including N:M sparsity, structured pruning, and mixed-precision quantization. The preliminary results are shown below.

\subsection{N:M Sparsity} 

Following DominoSearch~\citep{sun2021dominosearch}, we choose a mixed N:8 sparsity configuration. Instead of employing a uniform N value across all layers, we allow individual layers to have distinct N values while maintaining the overall parameter count constant. We use OWL to determine the optimal value of N for individual layers. The results are presented in Table~\ref{tab:NM}. It is evident that OWL consistently enhances performance compared to uniform N:M sparsity. Remarkably, in high sparsity scenarios like 3:8 and 2:8 sparsity, OWL demonstrates significant improvements with 2$\times$ and 8$\times$ perplexity reductions over the uniform baseline, respectively.

    
    

\begin{table*}[h]
\centering
\caption{Perplexity of mixed N:M sparsity (N refers to non-zero weights) with LLaMA-7B on WikiText.}
\vspace{-0.8em}
\resizebox{0.5\textwidth}{!}{
\begin{tabular}{lcc}
\toprule
\bf Method & \bf N:M Sparsity Structure  & \bf Perplexity  \\ %
\midrule
Wanda    & 4:8  & 8.57 \\
\gr  OWL \textit{w.} Wanda &  Mixed 4:8 & \textbf{8.55} \\
\midrule
Wanda  & 3:8  & 42.56 \\
\gr  OWL \textit{w.} Wanda  &   Mixed 3:8 & \textbf{21.49} \\
\midrule
Wanda  &2:8  & 2962.00 \\
\gr  OWL \textit{w.} Wanda  &  Mixed 2:8 & \textbf{331.37} \\
\bottomrule
\end{tabular}}
\label{tab:NM}
\end{table*}

\subsection{Structured Pruning} 

Instead of pruning individual weights, structured pruning involves the selective removal of an entire group of weights, which are more amenable to hardware speedup, including weight blocks, neurons, filters/channels, and attention heads~\citep{liu2023ten}. We adhere to the recent methodology introduced in LLM Pruner~\citep{ma2023llm}, wherein entire neurons and attention heads are removed. This action facilitates the direct acceleration of pruned LLMs on GPUs or TPUs. We replace the uniform layerwise sparsity used by LLM pruner with the non-uniform layerwise sparsity discovered by OWL. Table~\ref{tab:structured} again demonstrates that OWL achieves preferable performance compared to the uniform layerwise sparsity in the context of structured pruning.

\begin{table*}[h]
\small
\centering
\caption{Perplexity of Structure Pruning with LLaMA-7B on WikiText and PTB.}
\vspace{-0.8em}
\resizebox{0.7\textwidth}{!}{
\begin{tabular}{lcccccc}
\toprule
\bf Dataset & \bf Pruning Method & \bf Layerwise Sparsity & \bf 20\% & \bf 40\% & \bf 60\% & \bf 80\% \\
\midrule
WikiText & LLM Pruner  &  Uniform & 19.09 & 30.39 & 90.02 & 1228.17 \\
\gr WikiText  & LLM Pruner & OWL  & \bf  18.57& \bf 28.65& \bf 76.99&  \bf 321.64 \\
\midrule

PTB & LLM Pruner & Uniform & 29.51 & 66.90 & 192.06 & 1691.87 \\
 \gr   PTB   & LLM Pruner  & OWL  & \bf  28.82 & \bf 53.22 & \bf 150.16 & \bf 502.07 \\   
\bottomrule
\end{tabular}}
\label{tab:structured}
\end{table*}

\subsection{Mixed-Precision Quantization}

Leveraging our non-uniform layerwise sparsity, we can enhance mixed-precision quantization by assigning higher precision to layers exhibiting more outliers. Following the approach outlined in~\citep{tang2022mixed}, we utilize OWL to assign different bit precision to different layers, thereby facilitating a mixed-precision quantization strategy. Our baseline here involves selecting layers randomly and based on the $L_1$ norm of weights. It is evident that OWL also serves as a valuable indicator for selecting important layers in mixed-precision quantization, leading to improved quantization performance as shown in Table \ref{tab:ap_ablation}.

\begin{table*}[h]
\centering
\caption{Perplexity of mixed-precision quantization with LLaMA-7B on WikiText.}
\vspace{-0.8em}
\resizebox{0.48\textwidth}{!}{
\begin{tabular}{lcc}
\toprule
\bf Method & \bf Precision  & \bf Perplexity  \\ %
\midrule
Same Bit-width  & 2 Bit  & 104151.84 \\
Same Bit-width  & 3 Bit  & 25.82 \\
Same Bit-width  & 4 Bit  & 6.29 \\
\midrule
Select with random  & Mixed 3/4 Bit & 12.04 \\
Select with $L_1$ norm  & Mixed 3/4 Bit &  14.61\\
 \gr Select with OWL  & Mixed 3/4 Bit & \textbf{9.09} \\
\midrule
Select with random  & Mixed 2/3/4 Bit & 11455.54 \\
Select with $L_1$ norm  & Mixed 2/3/4 Bit &  13959.422\\
 \gr Select with OWL  & Mixed 2/3/4 Bit & \textbf{190.28} \\
\midrule
Select with random  & Mixed 2/4 Bit & 14817.12 \\
Select with $L_1$ norm  & Mixed 2/4 Bit & 33670.214 \\
 \gr Select with OWL  & Mixed 2/4 Bit & \textbf{7505.60} \\
\bottomrule
\end{tabular}}
\label{tab:ap_ablation}
\end{table*}


\clearpage

\section{Per-Block Vs. Per-Layer}
\label{app:per-block} 
As we mentioned before, we assign a distinct pruning ratio for each Transformer block instead of each layer. To provide more insights about this option, we provide the performance comparison between these two options and report their layerwise sparsity respectively. We report the sparsity ratio of 7 FC layers including \texttt{q\_proj}, \texttt{k\_proj}, \texttt{v\_proj}, \texttt{o\_proj}, \texttt{gate\_proj}, \texttt{down\_proj}, and \texttt{up\_proj} of layers 1, 2, 15, 30, and 31. We found that applying OWL in a per-layer manner leads to sub-optimal performance (Perplexity: 86.285 vs 24.55). We can see that applying OWL in a per-layer manner will lead to nearly uniform sparsity of certain layers across Transformer blocks, such as \texttt{v\_proj}, \texttt{gate\_proj}, and \texttt{up\_proj}, which might be undesirable for LLM pruning.

\begin{table*}[h]
\small
\centering
\caption{Layerwise sparsity of LLaMA-7B pruned with per-layer OWL (Sparsity: 70\%, Perplexity:86.285).}
\vspace{-0.8em}
\resizebox{0.7\textwidth}{!}{
\begin{tabular}{lccccccc}
\toprule
Layer & \texttt{q\_proj} & \texttt{k\_proj} & \texttt{v\_proj} & \texttt{o\_proj} & \texttt{gate\_proj} & \texttt{down\_proj} & \texttt{up\_proj}  \\
\midrule
1 & 0.613 & 0.613 & 0.613 & 0.613 & 0.613 & 0.613 & 0.613 \\
2 & 0.641 & 0.641 & 0.641 & 0.641 & 0.641 & 0.641 & 0.641 \\
5 & 0.608 & 0.608 & 0.608 & 0.608 & 0.608 & 0.608 & 0.608 \\
30 & 0.760 & 0.760 & 0.760 & 0.760 & 0.760 & 0.760 & 0.760 \\
31 & 0.721 & 0.721 & 0.721 & 0.721 & 0.721 & 0.721 & 0.721 \\
\bottomrule
\end{tabular}}
\label{tab:per-layer}
\end{table*}

\begin{table*}[h]
\small
\centering
\caption{Layerwise sparsity of LLaMA-7B pruned with per-block OWL (Sparsity: 70\%, Perplexity:24.55).}
\vspace{-0.8em}
\resizebox{0.7\textwidth}{!}{
\begin{tabular}{lccccccc}
\toprule
Layer & \texttt{q\_proj} & \texttt{k\_proj} & \texttt{v\_proj} & \texttt{o\_proj} & \texttt{gate\_proj} & \texttt{down\_proj} & \texttt{up\_proj}  \\
\midrule
1 & 0.639 & 0.638 & 0.691 & 0.598 & 0.710 & 0.696 & 0.713 \\
2 & 0.680 & 0.677 & 0.707 & 0.679 & 0.711 & 0.703 & 0.713 \\
15 & 0.698 & 0.693 & 0.710 & 0.706 & 0.709 & 0.703 & 0.712 \\
30 & 0.705 & 0.704 & 0.713 & 0.663 & 0.710 & 0.657 & 0.712 \\
31 & 0.702 & 0.701 & 0.712 & 0.670 & 0.710 & 0.621 & 0.711 \\

\bottomrule
\end{tabular}}
\label{tab:per-block}
\end{table*}

\section{Hyperparameters}
\label{app:hyper}
In this section, we share the hyperparameters used to reproduce the results in our experiments in Table \ref{tab:hyper}.

\begin{table}[h]
\centering
\caption{Hyperparameters used to obtain the results in this paper.}
\vspace{-0.8em}
\resizebox{0.23\textwidth}{!}{
\begin{tabular}{ccc}
\toprule
Model &  $\mathbf{M}$ & $\lambda$  \\ 
\toprule
 LLaMA-7B&  5  & 8\% \\
LLaMA-13B &  7 &  8\% \\
 LLaMA-30B&  5  & 8\% \\
LLaMA-65B & 5 &  20\% \\
OPT-6.7B & 10  & 8\%  \\
\bottomrule
\end{tabular}
}\vspace{-1em}
\label{tab:hyper}
\end{table}


\end{document}